\documentclass[hidelinks, letterpaper]{article} 
\usepackage{aaai21}                  
\usepackage{times}                   
\usepackage{helvet}                  
\usepackage{courier}                 
\usepackage[hyphens]{url}            
\usepackage{graphicx}                
\usepackage{natbib}                  
\usepackage{caption}                 
\usepackage[switch]{lineno}
\usepackage{amsfonts}
\usepackage{amsmath}
\usepackage{bm}
\usepackage{booktabs} 
\usepackage{color}

\newcommand{\ak}[1]{{\color{red}}}

\usepackage{subcaption}
\usepackage{listings}
\usepackage{float}
\newcommand{\comment}[1]{}
\newcommand{\tableitem}{~~\llap{\textbullet}~~}

\title{Simple Modifications to Improve Tabular Neural Networks}
\author{James Fiedler \\}
\affiliations{
    Indeed.com \\
    jfiedler@indeed.com
}

\begin{document}
\maketitle

\begin{abstract}
There is growing interest in neural network architectures for tabular data. Many general-purpose tabular deep learning models have been introduced recently, with performance sometimes rivaling gradient boosted decision trees (GBDTs). These recent models draw inspiration from various sources, including GBDTs, factorization machines, and neural networks from other application domains. Previous tabular neural networks are also drawn upon, but are possibly under-considered, especially models associated with specific tabular problems. This paper focuses on several such models, and proposes modifications for improving their performance. When modified, these models are shown to be competitive with leading general-purpose tabular models, including GBDTs.
\end{abstract}

\section{Introduction}

Many neural network architectures have been introduced lately as general-purpose tabular solutions. Some examples: TabNet \cite{arik2020tabnet}, TabTransformer \cite{huang2020tabtransformer}, NODE \cite{popov2019neural}, DNF-Net \cite{abutbul2020dnfnet}. The introduction of these and other models demonstrates increasing interest in the application of deep learning to tabular data. This is not due to a lack of solutions outside of deep learning. Gradient boosted decision trees (GBDTs) are very good general-purpose models, and in fact are frequently used by tabular deep learning models as both inspiration and the standard by which to measure performance.

Much of the interest in tabular deep learning models is due to advantages they already have over GBDTs or other established models. Possibly the biggest advantage is that neural networks have the potential to be end-to-end learners, removing the need for manual categorical encoding or other feature engineering. NN models also allow more training options than GBDTs. For example, they allow continual training on streaming data, and more generally allow adjustment of learned NN parameters by training on new data. Neural network models are also better suited to unsupervised pre-training; for example, see \cite{arik2020tabnet} and \cite{huang2020tabtransformer}.

Even before the recent rise of general-purpose tabular models, neural networks were proven effective in specific tabular-data problems, in particular, click-through-rate (CTR) and recommender systems. Many neural network models have used these two applications as primary motivation, including Wide\&Deep \cite{cheng2016wide}, Deep\&Cross \cite{wang2017deep}, Product Neural Networks (PNN) \cite{qu2016productbased}, xDeepFM \cite{Lian_2018}, AutoInt \cite{song_autoint_2019}, and FiBiNet \cite{huang2019fibinet}. Structurally, the techniques from these models are applicable to tabular problems more generally\footnote{Except, perhaps, the original version of PNN, which seems to have only considered categorical data}. However, apart from AutoInt, these models have not usually been featured in the performance comparisons of recent general tabular models.

It's reasonable to assume that models created intentionally to be general-purpose solutions would be superior to any of these more-specific models when evaluated across diverse tabular datasets. Some evidence supports this assumption. For example, \cite{gorishniy2021revisiting} compares AutoInt and a newer general tabular model, with AutoInt performing less well. On the other hand, there are also cases where models created for a specific purpose outperform newer general-purpose tabular models. That same paper also shows that ResNet \cite{he2015deep}, designed for image classification, can out-perform some general-purpose models. Hence existing tabular neural networks associated with a specific tabular problems might be worth considering when looking for more general solutions.

\begin{figure*}[ht]
\centering
\begin{subfigure}[b]{0.37\linewidth}
\includegraphics[width=\linewidth]{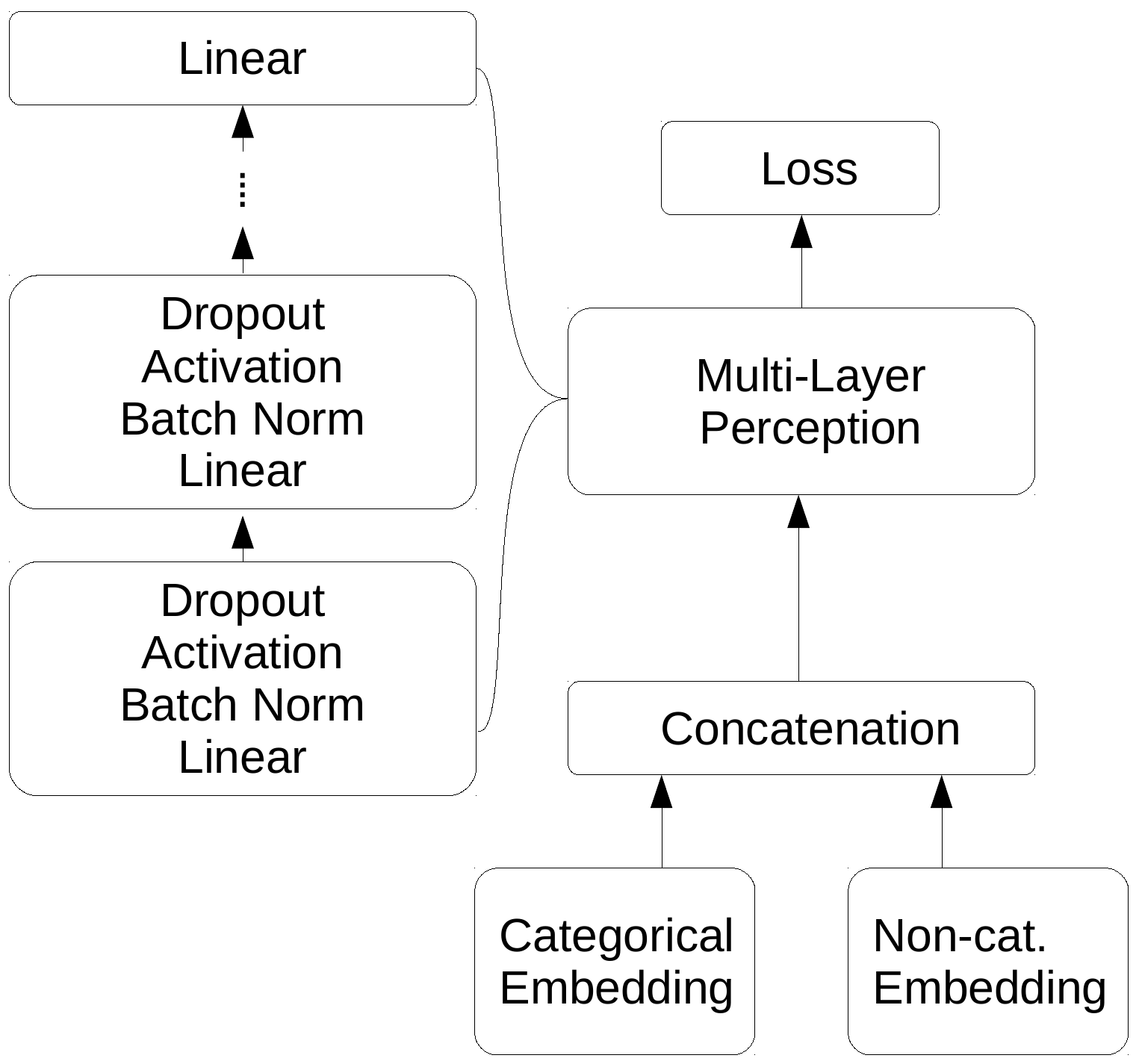} 
\caption{original MLP model}
\label{fig:mlp-model}
\end{subfigure}
\hspace*{1.0cm}
\begin{subfigure}[b]{0.40\linewidth}
\includegraphics[width=\linewidth]{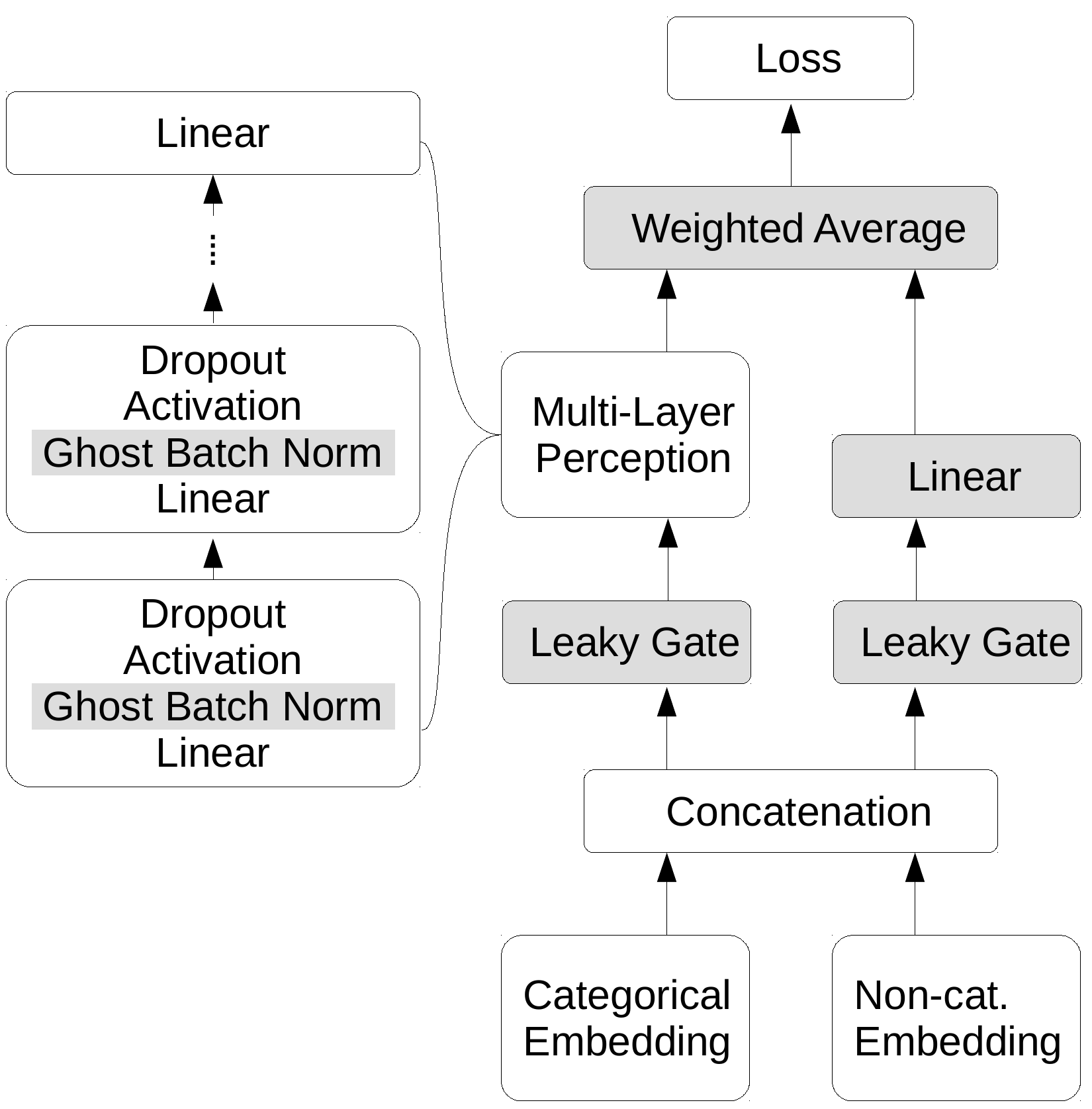} 
\caption{modified MLP model, ``MLP+"}
\label{fig:mlpplus-model}
\end{subfigure}
\caption{Original and modified versions of the MLP model. Modifications are given a gray background.}
\label{fig:mlp-models}
\end{figure*}

Looking in another direction for inspiration, multi-layer perceptrons (MLPs; see Figure \ref{fig:mlp-model})) are perhaps the simplest and most general-purpose NNs. Basic and modified versions are frequently included in model comparisons, sometimes performing surprisingly well (see results in Section \ref{sec:results} and related work in Section \ref{sec:related-work}). When they perform well, their simplicity make them an attractive option. Their simplicity also makes them an attractive option for implementing and testing new ideas, and interpreting their effect.

With this context in mind, the present paper considers the MLP, PNN, and AutoInt models, and proposes a few simple modifications to make them competitive with recent general-purpose tabular NNs. Results below indicate that this goal is achieved. In fact, the modified models appear to be superior to recently proposed models and perhaps even superior to GBDTs from LightGBM \cite{ke2017lightgbm}, on the datasets used in \cite{huang2020tabtransformer}.

The main contributions of this paper:
\begin{enumerate}
\item Several simple modifications that make MLP, PNN, and AutoInt models perform on par with, or better than, recent general-purpose tabular NNs and GBDTs
\item Model comparisons across a broad range of datasets showing the effectiveness of the proposed modifications 
\item A demonstration of how one modification in particular contributes to model interpretability
\end{enumerate}

The proposed modifications are likely useful more generally than shown here, including probably for some of the models compared against. One of the proposed modifications, Leaky Gates, is possibly new, though a simple construction; see Section \ref{sec:interpretability}. The other modifications have been used elsewhere, but have not been applied in this combination, nor to improve PNN, AutoInt, or similar models. Also, as far as I am aware this is the first time that modified versions of models like PNN and AutoInt have been shown to be competitive with newer general-purpose tabular NNs.

How this paper is organized: Section \ref{sec:modifications} describes the proposed modifications. Section \ref{sec:experiments} explains the experiments and provides results showing that the modified models are competitive with any of the comparison models, and possibly superior. Section \ref{sec:ablation} assesses the benefits of the modifications by comparing several ablated variations of the models. Section \ref{sec:interpretability} shows how one of the proposed modifications can help understand the model and interpret output. Section \ref{sec:related-work} discusses some very recent related work and how that work can inspire future investigation.

\section{Models and modifications}  \label{sec:modifications}

Ghost batch norm (GBN) will be used in all of the models in place of batch norm. This idea is taken from TabNet \cite{arik2020tabnet} and was originally proposed in \cite{hoffer2018train}. GBN allows the use of large batch sizes, but with batch norm parameters calculated on smaller sub-batches. One big motivation for using GBN here is to speed up training, but \cite{hoffer2018train} also showed that GBN improves generalization when using large batch sizes.

Leaky Gates will also be used in all of the models. These are a combination of two simple elements, an element-wise linear transformation followed by a LeakyReLU activation. Leaky Gates are possibly new, but their simplicity might make that unlikely. Also, more complicated constructions exist with the same basic effect. For more discussion about Leaky Gates and their effects, see Section \ref{sec:interpretability}.

\begin{figure*}[ht]
\centering
\begin{subfigure}[b]{0.22\linewidth}
\includegraphics[width=\linewidth]{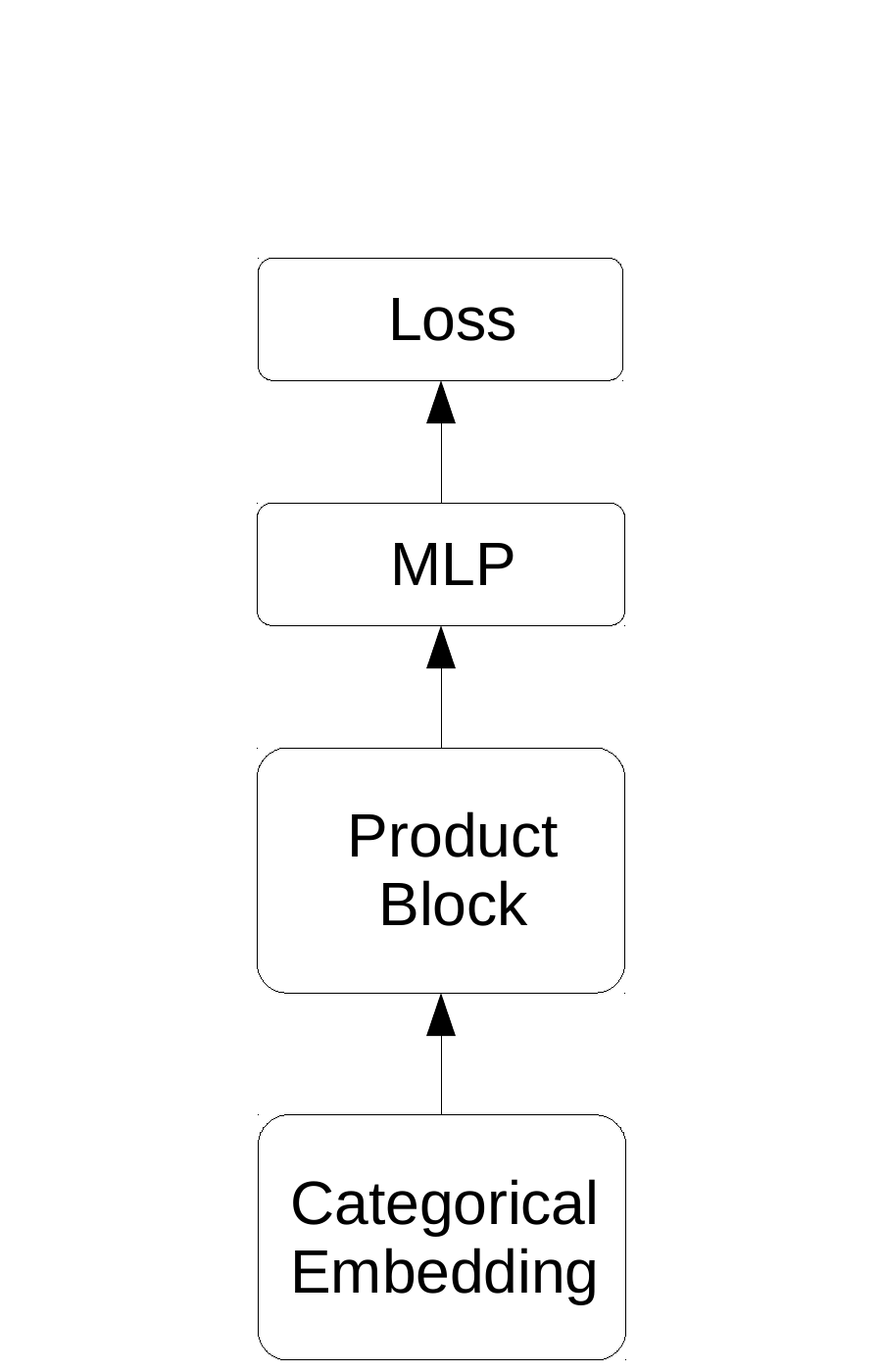} 
\caption{original PNN model\newline}
\label{fig:pnn-model-orig}
\end{subfigure}
\hspace*{1cm}
\begin{subfigure}[b]{0.22\linewidth}
\includegraphics[width=\linewidth]{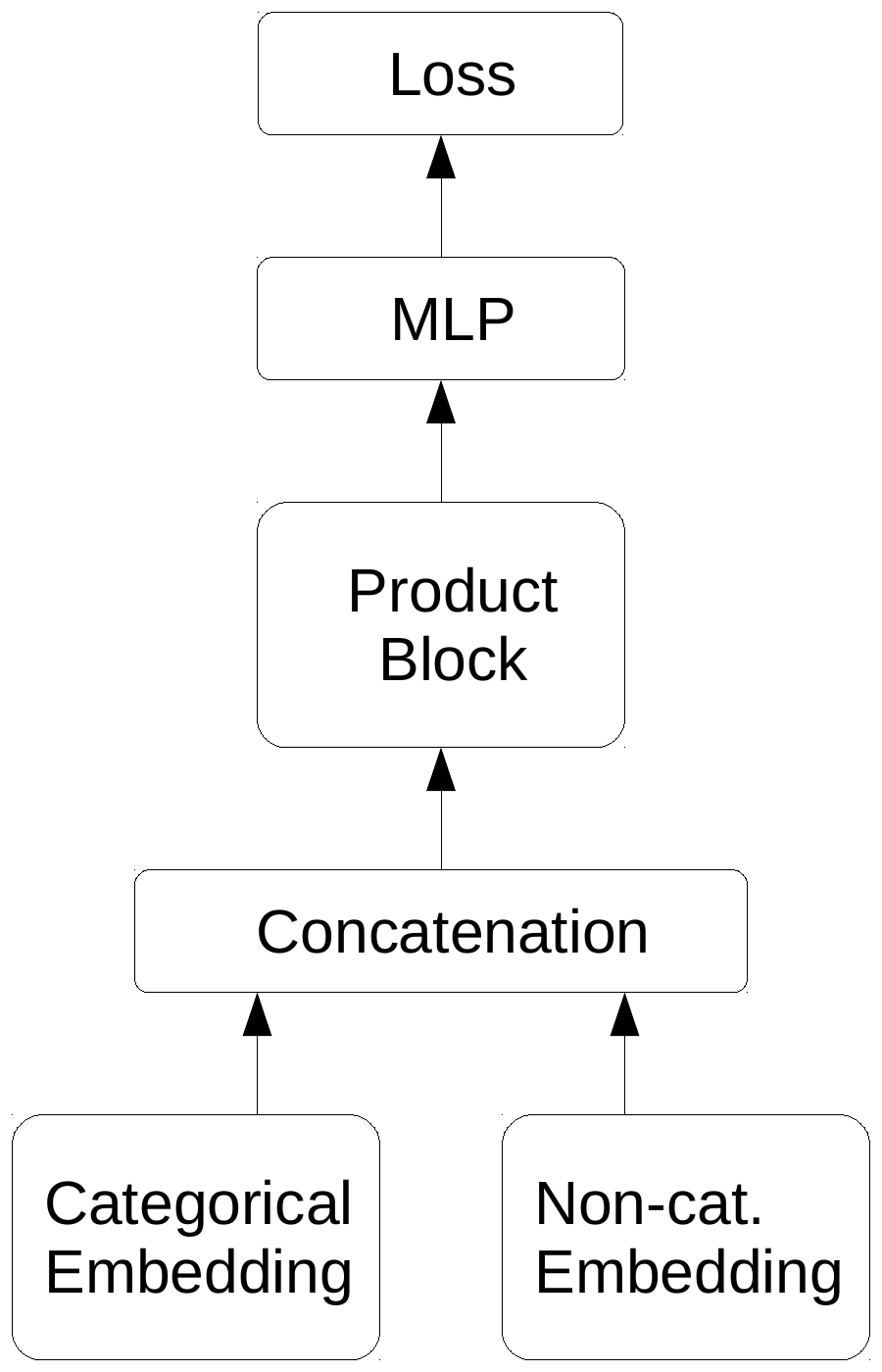} 
\caption{PNN model with embedded non-categorical values}
\label{fig:pnn-model}
\end{subfigure}
\hspace*{1.75cm}
\begin{subfigure}[b]{0.22\linewidth}
\includegraphics[width=\linewidth]{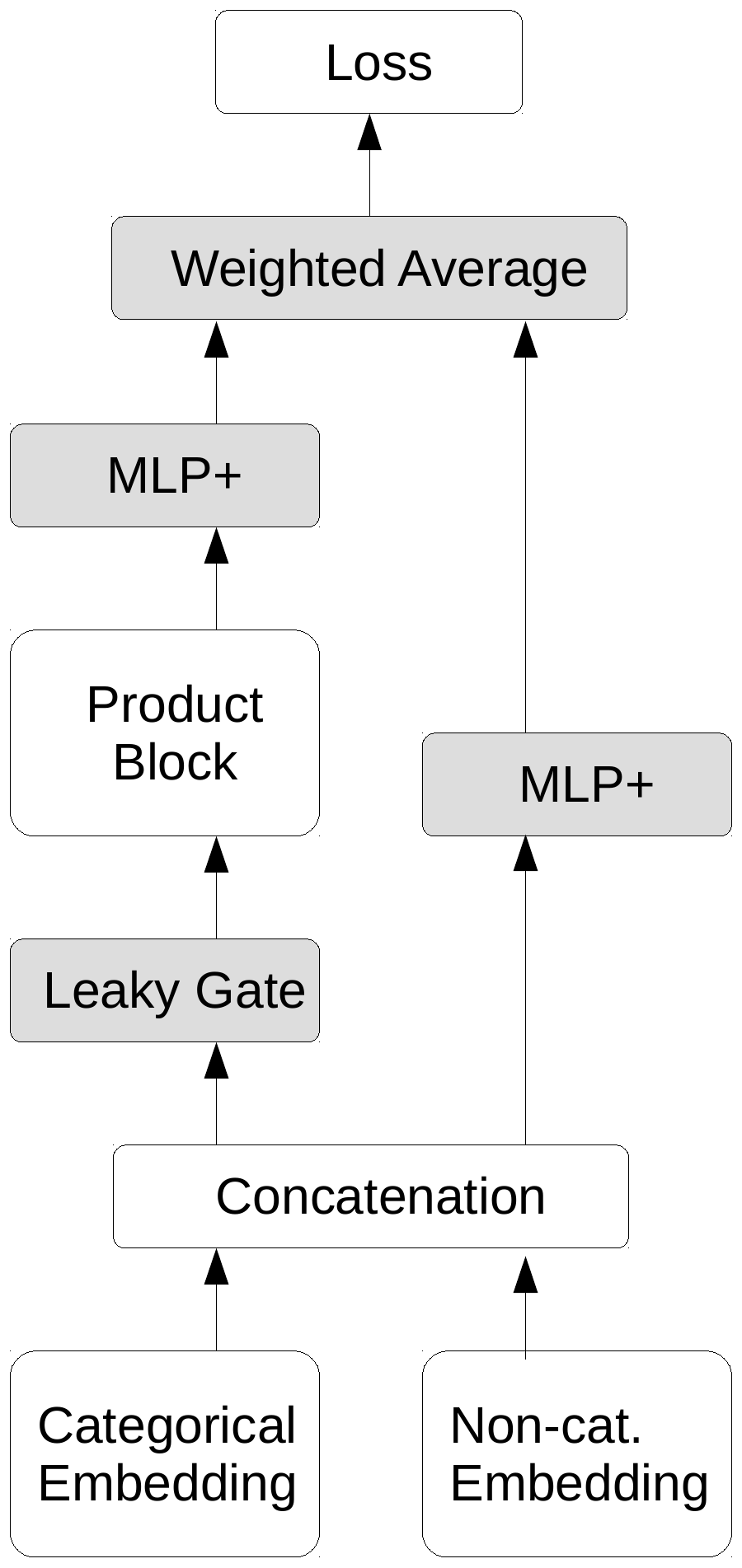} 
\caption{modified PNN model\newline}
\label{fig:pnnplus-model}
\end{subfigure}
\hspace*{0.7cm}
\caption{Original and modified versions of the PNN model. Modifications are given a gray background.}
\label{fig:pnn-models}
\end{figure*}

\subsection{MLP}

Figure \ref{fig:mlp-model} shows an MLP model before modification, and Figure \ref{fig:mlpplus-model} shows the MLP model afterwards, with a gray background for changed items.

Starting with the ``Multi-Layer Perceptron" sub-block, batch normalization is replaced with Ghost Batch Norm. Next, inspired by AutoGluon's MLP model \cite{erickson2020autogluontabular}, a linear skip layer is added to the right of the MLP sub-block. The skip layer is just a single fully-connected linear layer that creates a shorter path between embedding and the loss function. In the AutoGluon MLP, the skip layer's output is summed with the MLP sub-block's output. Here, a weighted average is taken, with the weight being a learned parameter. Finally, Leaky Gates are added before both the MLP sub-block and the linear skip layer.

As the bottom of Figure \ref{fig:mlp-models} shows, both categorical and numeric inputs are embedded. See Section \ref{sec:implementation-details} for details about the embeddings.

\subsubsection{Summary of changes to create MLP+}
\begin{itemize}
	\item Use Ghost Batch Norm in place of batch norm
	\item Add a linear skip layer
	\item Use a weighted average of the skip layer output and main layer outputs, with learned weight
	\item Add a Leaky Gate before both the MLP sub-block and the skip layer
\end{itemize}
Multiple versions of MLP will be referenced later. To reduce ambiguity, this new modified version will be called ``MLP+". Also, ``MLP+ block" will refer to the combination of Leaky Gates, MLP sub-block, skip layer, and weighted average.

\subsection{PNN}

The original PNN model \cite{qu2016productbased} has two main components after embedding, Figure \ref{fig:pnn-model-orig} and \ref{fig:pnn-model}. Embedded features first pass through an inner- or outer-product block (or both) to create features interactions. Then the interaction values pass through an MLP block.

In Figure \ref{fig:pnnplus-model}, the MLP component is modified as above to become an MLP+ block. A second MLP+ block is added beside these components to create a separate path from embedded input to output. This is similar to the skip layer in the MLP+ model, but here a block of multiple layers is added instead of a single layer. In a "two column" model like this, the two outputs are usually summed, but, as above, a weighted average is used, with a learned weight. A single Leaky Gate is then added before the product block (both MLP+ blocks also include Leaky Gates), and again both categorical and numeric inputs are embedded.

In the MLP+ above, the non-categorical embedding is structurally optional. In PNN and AutoInt it is required if all features are going to be allowed equivalent interaction. PNN and AutoInt could pass un-embedded values directly to the two MLP+ blocks, bypassing the interaction blocks. However, that would allow fewer interactions between categorical and non-categorical values, a limitation that seems sub-optimal in general.

Having non-categorical values bypass the interaction block is actually common in tabular models with this basic structure. The TabTransformer model \cite{huang2020tabtransformer} in the comparisons below is an example. Also, the PNN paper \cite{qu2016productbased} doesn't mention non-categorical values, so they likely either weren't allowed or were allowed but bypassed the interaction block.

\subsubsection{Summary of changes to PNN}
\begin{itemize}
	\item Add a Leaky Gate before the product block
	\item Change the MLP block to an MLP+ block
	\item Add a second MLP+ column
	\item Take a weighted average of the two MLP+ outputs, with learned weight
\end{itemize}

\subsection{AutoInt}

\begin{figure*}
\centering
\begin{subfigure}[b]{0.28\linewidth}
\includegraphics[width=\linewidth]{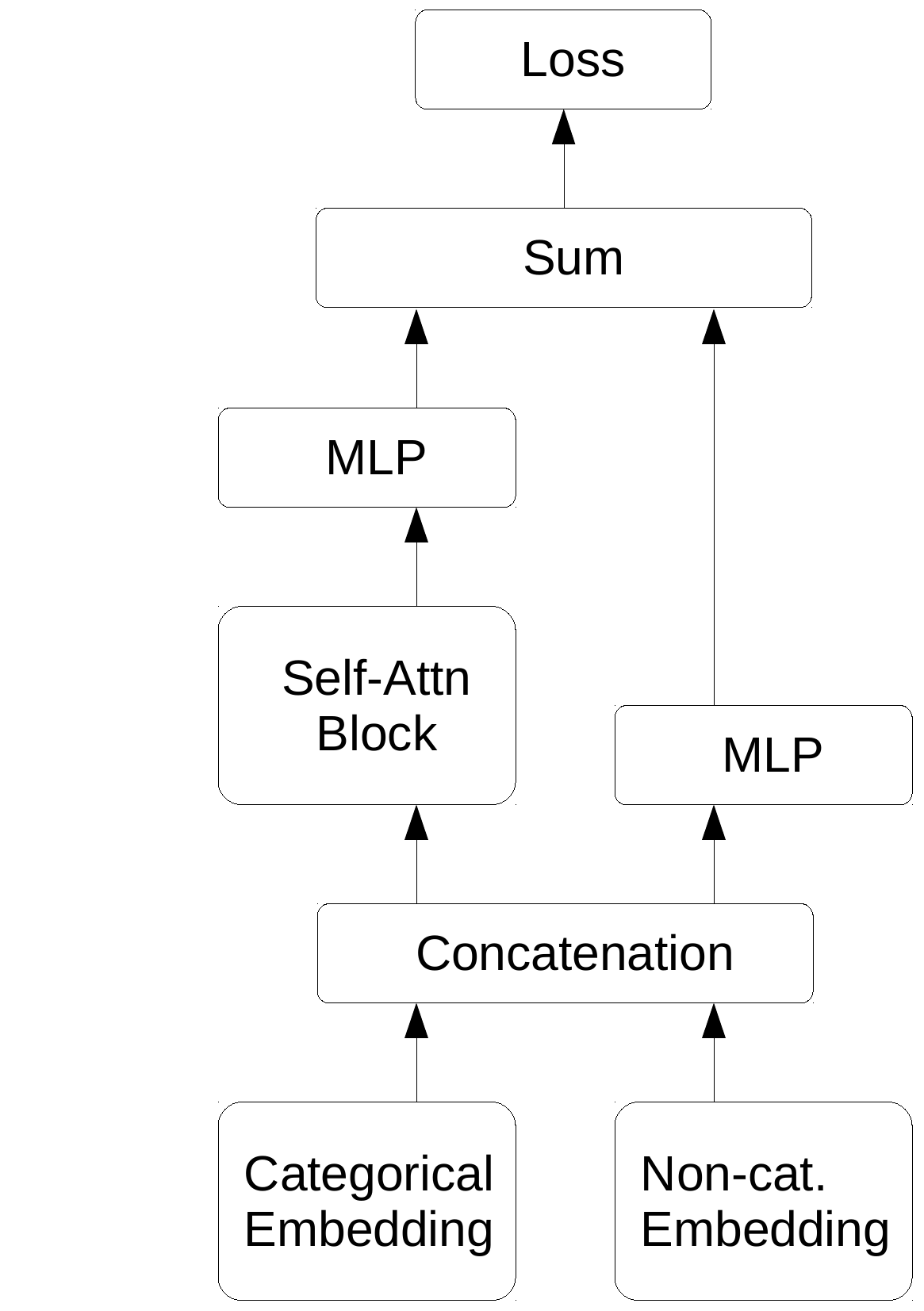} 
\caption{original AutoInt model}
\label{fig:autoint-model}
\end{subfigure}
\hspace*{1.5cm}
\begin{subfigure}[b]{0.28\linewidth}
\includegraphics[width=\linewidth]{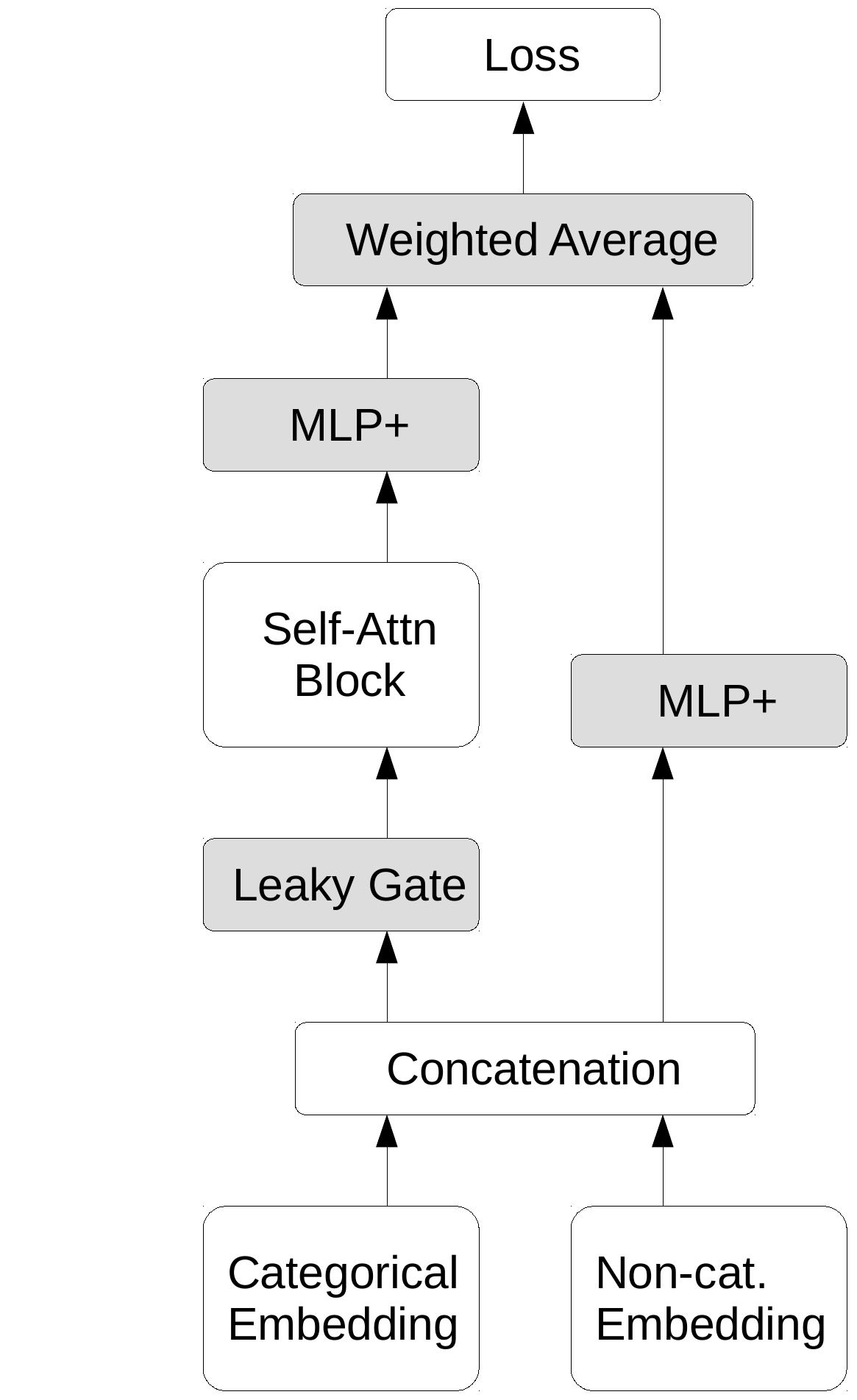} 
\caption{modified AutoInt model}
\label{fig:autointplus-model}
\end{subfigure}
\caption{Original and modified versions of the AutoInt model. Modifications are given a gray background.}
\label{fig:autoint-models}
\end{figure*}

In place of the product interaction block in PNN, AutoInt uses an interaction block with a multi-head self-attention mechanism \cite{song_autoint_2019}, inspired by \cite{vaswani2017attention}. Otherwise, the general structure of AutoInt and PNN are similar. Hence the modifications to the two-column version of AutoInt are nearly the same as to PNN, except a) AutoInt already had a second MLP column\footnote{More precisely, the original ``AutoInt" model has a one-column structure similar to the original PNN. The two-column version is actually called ``AutoInt+". However, only the two-column version is used here, so it is called ``AutoInt", even though that name is not entirely correct. Since the modified version has the two-column structure, it can be thought of as modifying the one-column structure by adding a new column, or modifying the two-column structure by making changes within the two columns.}, and b) AutoInt already used an embedding for non-categorical features.

\subsubsection{Summary of changes to AutoInt}
\begin{itemize}
	\item Add a Leaky Gate before the self-attention block
	\item Change the MLP blocks to MLP+ blocks
	\item Take a weighted average of the two MLP+ outputs, instead of a sum
\end{itemize}

\subsection{Other Candidate Models}

There are other tabular models that have the same basic structure as PNN or AutoInt and could possibly benefit from these modifications. For example, TabTransformer's structure is similar to PNN, and the structure of xDeepFM \cite{Lian_2018} is similar to AutoInt. Of course, any model that includes an MLP component could potentially benefit from the modifications in MLP+.

\subsection{Other implementation details}  \label{sec:implementation-details}

The implementations here use LeakyReLU \cite{maas2013rectifier} in places where ReLU would commonly be used. In particular, LeakyReLU is the activation used in the MLP+ model and MLP+ blocks within PNN and AutoInt.

All models use the same embedding scheme. Each categorical input field $f$ is given an embedding matrix $E_f$ of dimension $m \times c_f$ where $m$ is the embedding size and $c_f$ is the number of distinct values in the field. If $h$ is a one-hot embedding of the field's values, then for a single value $x_i$ the embedding $e_i$ is
\[
e_i = E_f \cdot h(x_i).
\]
Or, if the field values are encoded as contiguous integers $1, 2, \ldots, c_f$, then the embedding for $x_i$ is the $x_i^\text{th}$ column of $E_f$. The elements of $E_f$ are learned along with the other model parameters.

Non-categorical values will also be embedded, using the method in \cite{song_autoint_2019}. Each non-categorical field $f$ is given a column vector $V_f$. For a single value $x_i$, the embedding $e_i$ is
\[
e_i = x_i V_f.
\]
In other words, each individual value's embedding vector is just a scaling of the field's embedding vector.

The PNN and AutoInt models require the embedding size to be the same for all columns (both categorical and non-categorical). In the experiments below, the embedding size is a parameter to be optimized. MLP+ does not have the same structural requirement for uniform embedding size, and, as mentioned above, does not even structurally require non-categorical values to be embedded. However, MLP+ uses the same embedding scheme in the experiments for the sake of uniformity and to reduce the complexity of parameter search logic.

The MLP+, PNN, and AutoInt models are implemented in PyTorch \cite{pytorchpaper}. Code is available at \url{www.github.com/jrfiedler/xynn}

\section{Experiments}  \label{sec:experiments}

\subsection{Data}

The datasets are the same as in the TabTransformer paper \cite{huang2020tabtransformer}: 15 datasets from the UCI repository\footnote{\url{http://archive.ics.uci.edu/ml}}, the AutoML Challenge \cite{automlchallenges}, and Kaggle\footnote{\url{https://www.kaggle.com/datasets}}. For all datasets the task is binary classification. Dataset sizes range from 1,055 to 425,240 rows and 10 to 308 input features. More details are given in Appendix Table \ref{tab:data-details}.

Dataset splits are 5-fold cross-validation splits with 65/15/20\% for train/validation/holdout. The training dataset is the same for both validation and for the final holdout results. The holdout set is only used to determine the final model score. For models included in the TabTransformer paper, the 5-fold split for each dataset is the same across all models. The scores for those models are copied here. For model types introduced here, the same approach is used, but the original seed is not known, so two new seeds are used. Thus, for MLP+, PNN, and AutoInt a total of 10 fit models are used for each summary.

\subsection{Model parameters}

\subsubsection{Static parameters}

For each dataset, the batch size and ghost batch size are static for all models; see Appendix Table \ref{tab:data-details}. All models use cross-entropy loss, the Adam optimizer \cite{kingma2017adam} without weight decay, and PyTorch's StepLR scheduler to reduce the learning rate by multiplying by 0.95 at an interval of a chosen number of epochs.

\subsubsection{Parameter search}

Parameters were tuned using Bayesian hyper-parameter optimization (HPO) via the Optuna package \cite{akiba2019optuna}. Twenty rounds of optimization were used for each model on each split of the data. Optimal parameters, including number of training epochs, were determined by maximizing the validation set AUROC (area under the ROC curve) \cite{bradley1997use}, with an early stopping patience of 15 epochs. Models were allowed to use different parameters on different folds of the same dataset. For a complete list of the parameter search options, see Appendix Table \ref{tab:parameter-options}.


\subsection{Comparison models}
The results for MLP+, PNN, and AutoInt are compared against
\begin{itemize}
  \item Logistic regression (LR)
  \item Gradient boosted decision trees, specifically, LightGBM \cite{ke2017lightgbm}
  \item A simple MLP model, created by removing layers from the TabTransformer model (see \cite{huang2020tabtransformer}, \S 3.1, paragraph 1)
  \item A sparse MLP, based on \cite{morcos_one_2019} 
  \item TabTransformer \cite{huang2020tabtransformer}
  \item TabNet \cite{arik2020tabnet}
  \item Variational Information Bottleneck (VIB) \cite{alemi_deep_2016}
\end{itemize}
Results for these models are taken from \cite{huang2020tabtransformer}.

\subsection{Results}  \label{sec:results}

Table \ref{tab:overall-auroc-mean} gives the overall mean AUROC for each of the models across all datasets. For each individual dataset, the mean AUROC across cross-validation folds is given in Table \ref{tab:main-auroc-scores}. These tables show that MLP+, PNN, and AutoInt are competitive with any of the comparison models. In fact, the results suggest that these models outperform the other models. AutoInt, PNN, and MLP+ have the largest overall mean AUROC scores, and for 10 out of the 15 datasets AutoInt has the largest individual AUROC or is tied for the largest (rounded to 3 digits\footnote{Scores for all models other than MLP+, PNN, and AutoInt come from \cite{huang2020tabtransformer}, where they are rounded to 3 digits.}). In particular, AutoInt, PNN, and MLP+ seem to outperform the recently-introduced TabTransformer and TabNet models. The modified PNN's performance is possibly a little disappointing, though, since it contains two MLP+ blocks but overall performs only slightly better than the MLP+ model with a single MLP+ block.

\begin{table}[h]
\caption{Mean percent AUROC score for each model over all 15 datasets. Larger values are better. The models with proposed modifications are in bold.}
\centering
\label{tab:overall-auroc-mean}
\setlength{\tabcolsep}{4pt}
\scalebox{0.92}{
\begin{tabular}{lcc}
\toprule
                    & Mean \%  & Number best or \\
            Model   & AUROC    & tied for best  \\
\midrule
        \bf AutoInt & \bf 83.3 &  \bf 10  \\
            \bf PNN & \bf 83.1 &  \bf  3  \\
           \bf MLP+ & \bf 83.0 &  \bf  2  \\
           LightGBM &     82.9 &       5  \\
     TabTransformer &     82.8 &       1  \\
                MLP &     81.8 &       0  \\
         Sparse MLP &     81.4 &       0  \\
                VIB &     80.5 &       0  \\
Logistic Regression &     80.4 &       1  \\
             TabNet &     77.1 &       0  \\
\bottomrule
\end{tabular}
}
\end{table}

\begin{table*}[ht]
\caption{Percent AUROC scores on all datasets. Results for models LR through VIB are taken from \cite{huang2020tabtransformer}, and their values are the mean over 5 cross-validation splits. The last three columns are new, and their values are the mean over two random seeds and 5 cross-validation splits per seed (10 CV splits total). Larger AUROC values are better. Numbers in bold and the "Best Model" column are the highest or tied for highest result in each row, based on the 3 digits shown.}
\label{tab:main-auroc-scores}
\centering
\scalebox{0.87}{
\begin{tabular}{llcccccccccc}
	\toprule
                     &                               &           &  Light-   &        & Sparse &  Tab-   \\
	Dataset          &  Best Model                   &  LR       &  GBM      &  MLP   &  MLP   &  Trans.   & TabNet &  VIB   &  MLP+     &  PNN       &   AutoInt \\
	\midrule
	albert           &  LightGBM                     &     72.6 & \bf 76.3 &  74.0 &  74.1 &     75.7 &  70.5 &  73.7 &     74.1 &     74.1  &     73.9 \\
	hcdr\_main       &  LightGBM / AutoInt           &     74.7 & \bf 75.6 &  74.3 &  75.3 &     75.1 &  71.1 &  74.5 &     75.4 &     75.1  & \bf 75.6 \\
	dota2games       &  LR / MLP+ / AutoInt          & \bf 63.4 &     62.1 &  63.1 &  63.3 &     63.3 &  52.9 &  62.8 & \bf 63.4 &     63.3  & \bf 63.4 \\
	bank\_marketing  &  AutoInt                      &     91.1 &     93.3 &  92.9 &  92.6 &     93.4 &  88.5 &  92.0 &     93.7 &     93.5  & \bf 93.8 \\
	adult            &  AutoInt                      &     72.1 &     75.6 &  72.5 &  74.0 &     73.7 &  66.3 &  73.3 &     76.1 &     75.8  & \bf 76.2 \\
	1995\_income     &  AutoInt                      &     89.9 &     90.6 &  90.5 &  90.4 &     90.6 &  87.5 &  90.4 &     91.5 &     91.3  & \bf 91.9 \\
	online\_shoppers &  LightGBM                     &     90.8 & \bf 93.0 &  91.9 &  92.2 &     92.7 &  88.8 &  90.7 &     92.6 &     92.7  &     92.9 \\
	shrutime         &  AutoInt                      &     82.8 &     85.9 &  84.6 &  82.8 &     85.6 &  78.5 &  83.3 &     86.2 &     86.4  & \bf 86.6 \\
	blastchar        &  LGBM / MLP+ / PNN / AutoInt  &     84.4 & \bf 84.7 &  83.9 &  84.2 &     83.5 &  81.6 &  84.2 & \bf 84.7 & \bf 84.7  & \bf 84.7 \\
	philippine       &  TabTrans.                    &     72.5 &     81.2 &  82.1 &  76.4 & \bf 83.4 &  72.1 &  75.7 &     80.7 &     80.8  &     81.1 \\
	insurance\_co    &  PNN                          &     73.6 &     73.2 &  69.7 &  70.5 &     74.4 &  63.0 &  64.7 &     75.7 & \bf 76.6  &     75.3 \\
	spambase         &  LightGBM                     &     94.7 & \bf 98.7 &  98.4 &  98.0 &     98.5 &  97.5 &  98.3 &     98.3 &     98.3  &     98.6 \\
	jasmine          &  AutoInt                      &     84.6 &     86.2 &  85.1 &  85.6 &     85.3 &  81.6 &  84.7 &     85.9 &     86.2  & \bf 86.3 \\
	seismicbumps     &  PNN / AutoInt                &     74.9 &     75.6 &  73.5 &  69.9 &     75.1 &  70.1 &  68.1 &     74.3 & \bf 75.8  & \bf 75.8 \\
	qsar\_bio        &  AutoInt                      &     84.7 &     91.3 &  91.0 &  91.6 &     91.8 &  86.0 &  91.4 &     92.4 &     92.4  & \bf 92.9 \\
	\bottomrule
\end{tabular}
}
\end{table*}

It is always important to consider how results like these generalize. The datasets and experiment setup were taken from \cite{huang2020tabtransformer} specifically to avoid choices that potentially favored the new models over, e.g., TabTransformer or TabNet. However, there are other details which could impact the comparison; some may favor AutoInt, PNN, and MLP+, some may favor other models.
\begin{itemize}
\item Different parameter search space. For example, the possible dropout values for AutoInt, MLP+, and PNN were $0.0, 0.25, 0.50, 0.75$, but for the other deep learning models the dropout values were $0.0, 0.1, 0.2, \ldots, 0.5$.
\item Different random seeds were used to split the dataset for AutoInt, MLP+, and PNN vs the other models. Here, two random seeds were used and results averaged, to try to reduce any large difference due to seed. However, it's possible that the data splits here are more advantageous than the original, or vice versa.
\item For TabTransformer, in addition to the 20 HPO trials per model per data split, 50 initial rounds of hyper-parameter optimization were used on 5 datasets to determine the MLP layer sizes and some TabTransformer parameter values. For the new models, no initial optimization rounds were used. Instead, a few manual trials (far fewer than 50) were used to determine batch size and ghost batch size.
\item Different categorical embeddings were used in the comparison deep learning models vs the models here. Also, the new models have an additional non-categorical embedding. The view taken here is that these embeddings are part of the model, but it is also reasonable to consider embeddings separate from the rest of the model.
\item The general experiment setup (again, copied from the TabTransformer paper \cite{huang2020tabtransformer}), with 20 trials of Bayesian HPO per model per fold, with epoch chosen by validation AUROC instead of using a preset large number of epochs or iterations, with all tasks being binary classification, etc., might favor some models over others.
\end{itemize}

\section{Ablation}   \label{sec:ablation}

Table \ref{tab:ablation-mlp} shows performance of four models that represent a rough progression from a simple MLP to AutoInt. The first model is a baseline MLP from \cite{huang2020tabtransformer}. Second,  $\text{MLP}_{\text{FF}}$ is created from the MLP+ model by removing the skip and Leaky Gate layers (the subscript on $\text{MLP}_{\text{FF}}$ stands for skip=False, gate=False), i.e., the model depicted in Figure \ref{fig:mlp-model}, but with Ghost Batch Norm. Next is the MLP+ model. Finally the modified AutoInt model, from Figure \ref{fig:autointplus-model}, which includes the structure of MLP+ but adds a second column with Leaky Gate, self-attention block, and a second MLP+ block.

\begin{table*}[ht]
  \caption{Comparison between a baseline MLP from \cite{huang2020tabtransformer}, a basic $\text{MLP}_{\text{FF}}$, MLP+, and AutoInt. The $\text{MLP}_{\text{FF}}$ model is obtained from MLP+ by removing the Leaky Gates and skip layer (keeping the embeddings and Ghost Batch Norm; also see Table \ref{tab:ablation-all-scores}). The evaluation metric is AUROC in percentage. For $\text{MLP}_{\text{FF}}$, MLP+, and AutoInt the gain over the previous model is shown.}
  \centering
  \label{tab:ablation-mlp}
  \setlength{\tabcolsep}{4pt}
  \scalebox{0.87}{
  \begin{tabular}{lccrccrccr}
  \toprule
                   & Baseline MLP & \multicolumn{2}{c}{$\text{MLP}_{\text{FF}}$} & \multicolumn{2}{c}{MLP+} & \multicolumn{2}{c}{AutoInt} \\
           Dataset & AUROC & AUROC & Gain & AUROC & Gain & AUROC & Gain \\
  \midrule
            albert & 74.0 & 73.9 &    -0.1 & 74.1 & \bf 0.2 & 73.9 &    -0.2 \\
      1995\_income & 90.5 & 91.3 & \bf 0.8 & 91.5 & \bf 0.2 & 91.9 & \bf 0.4 \\ 
        dota2games & 63.1 & 63.3 & \bf 0.2 & 63.4 & \bf 0.1 & 63.4 &     0.0 \\
        hcdr\_main & 74.3 & 74.7 & \bf 0.4 & 75.4 & \bf 0.7 & 75.6 & \bf 0.2 \\
             adult & 72.5 & 76.0 & \bf 3.5 & 76.1 & \bf 0.1 & 76.2 & \bf 0.1 \\
   bank\_marketing & 92.9 & 93.5 & \bf 0.6 & 93.7 & \bf 0.2 & 93.8 & \bf 0.1 \\
         blastchar & 83.9 & 84.3 & \bf 0.4 & 84.7 & \bf 0.4 & 84.7 &     0.0 \\ 
     insurance\_co & 69.7 & 73.7 & \bf 4.0 & 75.7 & \bf 2.0 & 75.3 &    -0.4 \\
           jasmine & 85.1 & 85.7 & \bf 0.6 & 85.9 & \bf 0.2 & 86.3 & \bf 0.4 \\
  online\_shoppers & 91.9 & 92.2 & \bf 0.3 & 92.6 & \bf 0.4 & 92.9 & \bf 0.3 \\
        philippine & 82.1 & 80.2 &    -1.9 & 80.7 & \bf 0.5 & 81.1 & \bf 0.4 \\
         qsar\_bio & 91.0 & 92.6 & \bf 1.6 & 92.4 &    -0.2 & 92.9 & \bf 0.5 \\
      seismicbumps & 73.5 & 72.7 &    -0.8 & 74.3 & \bf 1.6 & 75.8 & \bf 1.5 \\
          shrutime & 84.6 & 85.8 & \bf 1.2 & 86.2 & \bf 0.4 & 86.6 & \bf 0.4 \\
          spambase & 98.4 & 97.9 &    -0.5 & 98.3 & \bf 0.4 & 98.6 & \bf 0.3 \\
  \midrule
              mean & 81.8 & 82.5 & \bf 0.7 & 83.0 & \bf 0.5 & 83.3 & \bf 0.3 \\
  \bottomrule
  \end{tabular}}
\end{table*}

In Table \ref{tab:ablation-mlp}, the overall mean AUROC is given, along with AUROC for each dataset and the gain over the \emph{previous} model, left to right. The overall mean improves for each successive model, and in most individual datasets the successive model improves on the previous model or is tied with it. However, the overall gain decreases with each successive model.

The difference in structure between the AutoInt and MLP+ models is large, as can be seen by comparing Figure \ref{fig:mlpplus-model} and Figure \ref{fig:autointplus-model}. There are also many differences between the baseline MLP and $\text{MLP}_{\text{FF}}$:
\begin{itemize}
\item The baseline MLP from \cite{huang2020tabtransformer} uses a SELU activation \cite{klambauer2017self}, $\text{MLP}_{\text{FF}}$ uses LeakyReLU.
\item The baseline MLP used batch norm, $\text{MLP}_{\text{FF}}$ uses Ghost Batch Norm.
\item The search space for linear layer sizes and dropout rate was different in \cite{huang2020tabtransformer} vs here. See Appendix Table \ref{tab:parameter-options} for details on the search space.
\end{itemize}
All of these implementation details for $\text{MLP}_{\text{FF}}$ also apply to the modified MLP+, PNN, and AutoInt models.

Table \ref{tab:ablation-summary} summarizes another ablation, from the fully-modified models to versions that remove the Leaky Gates and the skip layer. For all three models, the fully-modified version is best. For MLP+, dropping the skip layer is better than dropping both the skip layer and Leaky Gate, as might be expected. For PNN, though, that is the worst option, and for AutoInt it is equal to dropping both the skip layer and Leaky Gate.

Appendix Table \ref{tab:ablation-all-scores} includes scores for all three versions for all datasets.

\begin{table}[ht]
  \caption{Percent AUROC scores for the new models under three scenarios. 1) Same as reported above, using Leaky Gates and MLP skip layers, 2) removing Leaky Gates but keeping MLP skip layers, and 3) removing both Leaky Gates and MLP skip layers.}
	\label{tab:ablation-summary}
	\centering
		\begin{tabular}{rrrr}
			\toprule
                     & \multicolumn{3}{c}{Mean \% AUROC} \\
                     & skip=T & skip=T & skip=F  \\
                     & gate=T & gate=F & gate=F  \\
    			\midrule
               MLP+  &   83.0 &   82.8 &   82.5  \\
               PNN   &   83.1 &   82.6 &   82.9  \\
             AutoInt &   83.3 &   82.6 &   82.6  \\
			\bottomrule
	\end{tabular}
\end{table}

Many more versions of ablation could be carried out to test more subsets of the proposed modifications. However, the general message from the current ablation results is that the proposed modifications improve the models and work well together.

\section{Interpreting Leaky Gate Output}  \label{sec:interpretability}

The Leaky Gate is a combination of an element-wise linear transformation and a LeakyReLU. The first layer will scale, translate, and/or flip each column's values independently of the other columns. The LeakyReLU will let any positive value through unchanged and will squeeze any negative values almost to zero. In other words if $w_i$ and $b_i$ are the linear layer's parameters for the $i^\text{th}$ column, then the gate's effect for column $i$ is
\[
g_i(x_i) = \begin{cases}
w_i x_i + b_i & \text{if } w_i x_i + b_i > 0 \\
\approx 0 & \text{otherwise} \\
\end{cases}.
\]
In the first case, the value ``passes through", i.e., the value gets through intact, and in the second case it ``leaks through"\footnote{This distinction depends on the value of $w_i$ being larger than the LeakyReLU's slope for negative values, but in manually inspected examples these distinctions hold.}. Depending on the signs of $w_i$ and $b_i$, the gate determines how the column's values are partitioned between passing through the gate or leaking through, Table \ref{tab:passage-case}.

\begin{table}[ht]
\caption{
	How a column's values $x_i$ are partitions by a Leaky Gate, depending on $w_i$ and $b_i$, the linear layer's parameters for the $i^\text{th}$ column.
}
\label{tab:passage-case}
\centering
\begin{tabular}{ccc}
\toprule
case & $x_i$ passes through & $x_i$ leaks through \\
\midrule
$w_i > 0$             & $(-b_i / w_i, +\infty)$ & $(-\infty, -b_i/w_i]$ \\
$w_i < 0$             & $(-\infty, -b_i/w_i)$   & $[-b_i / w_i, +\infty)$ \\
$w_i = 0; b_i > 0$    & $(-\infty, +\infty)$    & $\emptyset$ \\
$w_i = 0; b_i \leq 0$ & $\emptyset$             & $(-\infty, +\infty)$ \\
\bottomrule
\end{tabular}
\end{table}

This partitioning behavior of the Leaky Gate is in fact the motivation for it. The Leaky Gate is intended to act as a simple filter or mask with potentially different behavior for each column, and where masking depends on each individual value in the column. The basic effect and intention are similar to the masking in TabNet \cite{arik2020tabnet}. One nice consequence of the filtering is in how it can potentially aid interpretation of the model in various ways, such as helping with a separate feature selection step or case-wise interpretation of results.

Here is an example involving feature selection. An MLP+ model is fit on the spambase dataset from Section \ref{sec:experiments}. This dataset has no categorical features. For the experiments, all features were embedded in a higher-dimensional space, but for this example, the features are not embedded. Otherwise, all aspects of the fitting process are the same as for the experiments: twenty rounds of Bayesian hyper-parameter optimization on the same parameter space, etc.

In the MLP+ model there are two Leaky Gates, one before the MLP block, and one before the skip layer, Figure \ref{fig:mlpplus-model}. since the data is not embedded, each column of input into the Leaky Gate is the original data column. For each of the Leaky Gates and for each column, we can count how many values passed through the gate vs leaked through by looking at the sign of the value either directly before the LeakyReLU or directly after. If the value is positive it passed through, otherwise it didn't.

\begin{figure}[h]
\centering
\includegraphics[width=\linewidth,trim={1cm 1.7cm 1.5cm 1.5cm},clip]{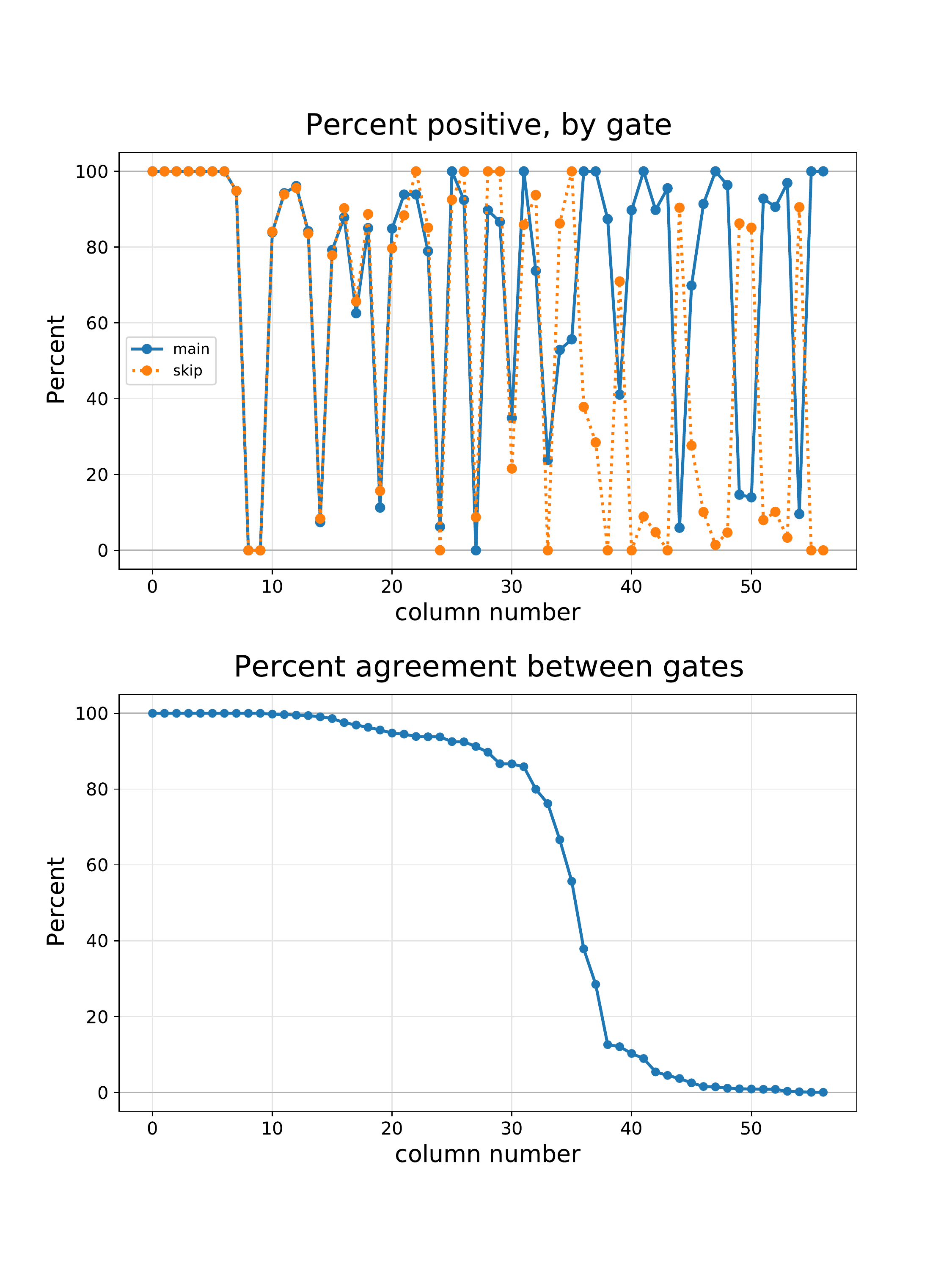}
\caption{Top: For each column and for each Leaky Gate, the percent of rows in training + validation with positive output value. Bottom: The percent of rows in which the two gates agree, i.e., both outputs are positive or both are non-positive.}
\label{fig:leaky-gate-output}
\end{figure}

Figure \ref{fig:leaky-gate-output} shows, column by column, the percent of rows that were positive after each Leaky Gate. The columns are sorted by the percent of rows on which the two Leaky Gates agree on positive-vs-not-positive (they agree if both are positive or both non-positive, otherwise they disagree). In the bottom plot we see agreement ranges from 100\% on the left to 0\% on the right. In fact, there is perfect agreement for ten columns, and there is perfect disagreement for two columns. Table \ref{tab:gate-passage} summarizes a few of the same values and might be easier to interpret.

From the top plot of Figure \ref{fig:leaky-gate-output} or from Table \ref{tab:gate-passage}, columns numbered 8 and 9 have no positive values after either gate. This suggests that those columns have little to no impact on the output of the MLP+. Hence they are excellent candidates for dropping. In fact, after dropping them and refitting, AUROC improved from 0.977 to 0.979. This AUROC is not as good as obtained when the columns were embedded, Table \ref{tab:main-auroc-scores}, but regardless, the example demonstrates the potential of the technique. The same technique could possibly be applied to the embedded columns, to drop unnecessary embedding dimensions.

\begin{table}[t]
\caption{
	For each column in spambase and for each Leaky Gate in MLP+, the percent of rows in training + validation sets with positive output value. For this example the raw values are used, i.e., not embedded.
}
\label{tab:gate-passage}
\centering
\scalebox{0.90}{
\begin{tabular}{rrrr}
\toprule
       & \multicolumn{2}{c}{\% positive}          \\
\cmidrule(lr){2-3}
column & main gate   & skip gate   & \% agreement \\
\midrule
     0 &       100.0 &       100.0 &        100.0 \\
     1 &       100.0 &       100.0 &        100.0 \\
     2 &       100.0 &       100.0 &        100.0 \\
     3 &       100.0 &       100.0 &        100.0 \\
     4 &       100.0 &       100.0 &        100.0 \\
     5 &       100.0 &       100.0 &        100.0 \\
     6 &       100.0 &       100.0 &        100.0 \\
     7 &        94.8 &        94.8 &        100.0 \\
     8 &         0.0 &         0.0 &        100.0 \\
     9 &         0.0 &         0.0 &        100.0 \\
    10 &        83.9 &        84.1 &         99.8 \\
    \vdots  \\
    54 &         9.6 &        90.5 &          0.1 \\
    55 &       100.0 &         0.0 &          0.0 \\
    56 &       100.0 &         0.0 &          0.0 \\
\bottomrule
\end{tabular}
}
\end{table}

\section{Related work}   \label{sec:related-work}

\subsubsection{MLPs with regularization}

A very recent article \cite{kadra2021regularization} showed that MLPs with a ``cocktail" of regularization strategies can obtain excellent performance. In that paper the MLP architecture was fixed and optimization focused on selecting regularization techniques from five categories:

\begin{enumerate}
\item Weight decay, e.g., $\ell_1$, $\ell_2$ regularization
\item Data augmentation, e.g., Cut-Out \cite{devries2017improved} and Mix-Up \cite{zhang2018mixup}
\item Model averaging, e.g., dropout and explicit average of models
\item Structural and linearization, e.g., skip layers
\item Implicit, e.g., batch normalization
\end{enumerate}

The approach was tested on a large collection of datasets and obtained better overall performance than GBDTs from XGBoost \cite{chen2016xgboost} and models such as TabNet, Neural Oblivious Decision Ensembles (NODE), \cite{popov2019neural}, and DNF-Net \cite{abutbul2020dnfnet}.

There are a lot of similarities between those results and the results here. The MLP+, PNN, and AutoInt models have techniques from 3 of the 5 regularization categories: dropout, skip layers, batch normalization (via Ghost Batch Norm)\footnote{Actually, \cite{dimitriou2020new} shows that the regularization effect of GBN is slightly different from batch norm, and GBN gives superior performance in their experiments.}, and explicit averaging of sub-components. The results here showed that the modified MLP+ outperformed GBDTs (but using LightGBM instead of XGBoost). The results here go one step further and show that the modifications used for MLP+ can also improve other tabular neural network models.

The main lesson from \cite{kadra2021regularization} for the current approach seems to be that more regularization strategies should be tested in combination with the techniques already used here. This is an excellent option for future investigation.

\subsubsection{More GBDT comparisons}

Similarly, \cite{shwartzziv2021tabular} compared XGBoost GBDTs against TabNet, NODE, DNF-Net, and 1D-CNN, a model introduced in a Kaggle competition \cite{baosenguo}. The datasets were comprised of nine datasets taken from the TabNet, NODE, and DNF-Net papers, three datasets from each paper, and two datasets that did not appear in any of the papers (shrutime and blastchar datasets, also used in the current paper). The results show that GBDTs generally outperform any single recent neural network. Another interesting finding is that the recent neural networks generally perform much better on datasets from their own papers. In other words, the recent tabular neural networks that they look at do not seem to generalize well. One lesson to take away is that the modified models introduced here should be tested on a variety of additional datasets. 

\subsubsection{More GBDT comparisons; another transformer model}

Another recent paper, \cite{gorishniy2021revisiting}, compares GBDTs and tabular neural networks, and argues that GBDTs and neural networks perform well on different problems. It uses GBDTs from XGBoost and CatBoost \cite{prokhorenkova2018catboost}, and TabNet, NODE, AutoInt (not the modified version used in the current paper) and other neural network models. The neural network models performed better when data was ``homogeneous", when the concepts measured in the data were the same or very similar from feature to feature. For example, images with each pixel location as a different field would be homogeneous. GBDTs performed better when features were ``heterogeneous". For example, housing data that records number of rooms, number of bedrooms, square footage, etc.

They also found that ResNet, though not originally intended as a tabular model, outperforms recent models. Finally, they introduced a new transformer-based model called FT-Transformer which outperformed ResNet and the other neural networks, but generally only outperformed GBDTs on homogeneous datasets.

The result that recent tabular neural networks perform less well than GBDTs is supported by \cite{huang2020tabtransformer}, which is the source of the current paper's comparison data. This can be seen in Table \ref{tab:overall-auroc-mean} where GBDTs outperform all of the models not in bold, judging by overall mean AUROC. Of course, Table \ref{tab:overall-auroc-mean} also suggests that the modified models introduced here do perform better than GBDTs, if the results here generalize. It's also not the case that the models introduced here generally perform worse on heterogeneous data. Several of the individual datasets where they perform best are heterogeneous, e.g., 1995\_income, shrutime, insurance\_co, qsar\_bio (see Table \ref{tab:main-auroc-scores}).

\section{Conclusion}

This paper proposes several modifications for tabular neural networks. These modifications are likely useful for many models, but three existing models are used for demonstration: MLP, PNN, and AutoInt. Experiments on 15 datasets and comparisons against seven other models show that the modified models are competitive with or outperform recent tabular neural networks and GBDTs from LightGBM. Investigating model variations with fewer modifications shows that the proposed combination works well.

The Leaky Gate proposed here is perhaps new, though a simple construction. Leaky Gates are potentially useful for interpretation of the model in various ways. In particular, an example showed that Leaky Gate output could be used to determine features that could safely be dropped from the input.

\begin{quote}
\begin{small}
\bibliography{reference}
\end{small}
\end{quote}

\newpage
\clearpage

\appendix

\begin{table*}
\caption{Parameter values used during optimization.}
\label{tab:parameter-options}
\centering
\scalebox{0.95}{
\begin{tabular}{lll}
\toprule
All models & PNN Product & AutoInt Self-Attention \\
\midrule
\tableitem MLP layer sizes             & \tableitem Product type: inner, outer, or both & \tableitem Embedding size: 8, 16, 32 \\
\phantom{\tableitem}\tableitem (256, 192, 128, 64)     &  \tableitem Output size: 20, 40, 80, 120 & \tableitem Number layers: 3, 4 \\
\phantom{\tableitem}\tableitem (512, 256, 128, 64)     & & \tableitem Number heads: 2, 3 \\
\phantom{\tableitem}\tableitem (512, 256, 128, 64, 32) & & \tableitem Dropout: 0.0, 0.1  \\
\phantom{\tableitem}\tableitem (1024, 512, 256, 128)   & & \tableitem Activation: None, LeakyReLU \\
\tableitem Dropout: 0.0, 0.25, 0.50, 0.75 & & \tableitem Use residual: True, False \\
\tableitem Learning rates: 0.1, 0.01, 0.001   \\
\tableitem Step sizes for learning rate \\ 
\phantom{\tableitem}scheduler: 10, 15, 20 epochs         \\
\bottomrule
\end{tabular}
}
\end{table*}

\begin{table*}
\caption{
	Dataset sizes and other details. For the dota2games and seismicbumps datasets, several constant features are dropped when using the data.
}
\label{tab:data-details}
\centering
	\begin{tabular}{lrrrrrrr}
		\toprule
                         &            &          & Positive & Batch & Ghost \\
		Dataset          & Datapoints & Features & Class \% & Size  & Batch \\
		\midrule
		albert           &  $425,240$ &       78 &     50.0 &  2048 &  64 \\
		hcdr\_main       &  $307,511$ &      120 &      8.1 &  1024 & 128 \\
		dota2games       &   $92,650$ &      116 &     52.7 &  1024 & 256 \\
		bank\_marketing  &   $45,211$ &       16 &     11.7 &  2048 &  16 \\
		adult            &   $34,190$ &       24 &     85.4 &  2048 &  32 \\
		1995\_income     &   $32,561$ &       14 &     24.1 &  2048 &   8 \\
		online\_shoppers &   $12,330$ &       17 &     15.5 &  2048 &   8 \\
		shrutime         &   $10,000$ &       10 &     20.4 &  2048 &   8 \\
		blastchar        &    $7,043$ &       19 &     26.5 &  2047 &   8 \\
		philippine       &    $5,832$ &      308 &     50.0 &   512 &   8 \\
		insurance\_co    &    $5,822$ &       85 &      6.0 &  1024 &   8 \\
		spambase         &    $4,601$ &       57 &     39.4 &  1024 &   8 \\
		jasmine          &    $2,984$ &      144 &     50.0 &   512 &   8 \\
		seismicbumps     &    $2,583$ &       18 &      6.6 &  2048 &   8 \\
		qsar\_bio        &    $1,055$ &       41 &     33.7 &  2048 &   8 \\
		\bottomrule
\end{tabular}
\end{table*}

\begin{table*}
\caption{Mean percent AUROC score on all datasets, plus or minus the standard deviation. For logistic regression through VIB, the values are from \cite{huang2020tabtransformer}, and the mean is taken over 5 cross-validation splits. For MLP+, PNN, and AutoInt, the mean is taken over 2 sets of 5 cross-validation splits.}
\label{tab:supervised-result-auroc1}
\centering
\scalebox{0.78}{
\begin{tabular}{lllllllllll}
\toprule
                 & Logistic       &                &                &                &  Tab- \\
Dataset          & Regression     &  LightGBM      &  MLP           &  Sparse MLP    &  Transformer   &  TabNet        &  VIB           & MLP+           & PNN            & AutoInt \\
\midrule
albert           & 72.6 $\pm$ 0.1 & 76.3 $\pm$ 0.1 & 74.0 $\pm$ 0.1 & 74.1 $\pm$ 0.1 & 75.7 $\pm$ 0.2 & 70.5 $\pm$ 0.5 & 73.7 $\pm$ 0.1 & 74.1 $\pm$ 0.2 & 74.1 $\pm$ 0.2 & 73.9 $\pm$ 0.4 \\
hcdr\_main       & 74.7 $\pm$ 0.4 & 75.6 $\pm$ 0.4 & 74.3 $\pm$ 0.4 & 75.3 $\pm$ 0.4 & 75.1 $\pm$ 0.4 & 71.1 $\pm$ 0.6 & 74.5 $\pm$ 0.5 & 75.4 $\pm$ 0.5 & 75.1 $\pm$ 0.4 & 75.6 $\pm$ 0.6 \\
dota2games       & 63.4 $\pm$ 0.3 & 62.1 $\pm$ 0.4 & 63.1 $\pm$ 0.2 & 63.3 $\pm$ 0.4 & 63.3 $\pm$ 0.2 & 52.9 $\pm$ 2.5 & 62.8 $\pm$ 0.3 & 63.4 $\pm$ 0.3 & 63.3 $\pm$ 0.3 & 63.4 $\pm$ 0.4 \\
bank\_marketing  & 91.1 $\pm$ 0.5 & 93.3 $\pm$ 0.3 & 92.9 $\pm$ 0.3 & 92.6 $\pm$ 0.7 & 93.4 $\pm$ 0.4 & 88.5 $\pm$ 1.7 & 92.0 $\pm$ 0.5 & 93.7 $\pm$ 0.3 & 93.5 $\pm$ 0.3 & 93.8 $\pm$ 0.3 \\
adult            & 72.1 $\pm$ 1.0 & 75.6 $\pm$ 1.1 & 72.5 $\pm$ 1.0 & 74.0 $\pm$ 0.7 & 73.7 $\pm$ 0.9 & 66.3 $\pm$ 1.6 & 73.3 $\pm$ 0.9 & 76.1 $\pm$ 0.9 & 75.8 $\pm$ 0.8 & 76.2 $\pm$ 0.9 \\
1995\_income     & 89.9 $\pm$ 0.2 & 90.6 $\pm$ 0.2 & 90.5 $\pm$ 0.3 & 90.4 $\pm$ 0.4 & 90.6 $\pm$ 0.3 & 87.5 $\pm$ 0.6 & 90.4 $\pm$ 0.3 & 91.5 $\pm$ 0.4 & 91.3 $\pm$ 0.4 & 91.9 $\pm$ 0.5 \\
online\_shoppers & 90.8 $\pm$ 1.5 & 93.0 $\pm$ 0.8 & 91.9 $\pm$ 1.0 & 92.2 $\pm$ 1.1 & 92.7 $\pm$ 1.0 & 88.8 $\pm$ 2.0 & 90.7 $\pm$ 1.2 & 92.6 $\pm$ 0.6 & 92.7 $\pm$ 0.7 & 92.9 $\pm$ 0.5 \\
shrutime         & 82.8 $\pm$ 1.3 & 85.9 $\pm$ 0.9 & 84.6 $\pm$ 1.3 & 82.8 $\pm$ 0.7 & 85.6 $\pm$ 0.5 & 78.5 $\pm$ 2.4 & 83.3 $\pm$ 1.1 & 86.2 $\pm$ 0.9 & 86.4 $\pm$ 0.9 & 86.6 $\pm$ 1.0 \\
blastchar        & 84.4 $\pm$ 1.0 & 84.7 $\pm$ 1.6 & 83.9 $\pm$ 1.0 & 84.2 $\pm$ 1.5 & 83.5 $\pm$ 1.4 & 81.6 $\pm$ 1.4 & 84.2 $\pm$ 1.2 & 84.7 $\pm$ 0.9 & 84.7 $\pm$ 1.0 & 84.7 $\pm$ 1.0 \\
philippine       & 72.5 $\pm$ 2.2 & 81.2 $\pm$ 1.3 & 82.1 $\pm$ 2.0 & 76.4 $\pm$ 1.8 & 83.4 $\pm$ 1.8 & 72.1 $\pm$ 0.8 & 75.7 $\pm$ 1.8 & 80.7 $\pm$ 1.4 & 80.8 $\pm$ 1.3 & 81.1 $\pm$ 1.8 \\
insurance\_co    & 73.6 $\pm$ 2.3 & 73.2 $\pm$ 2.2 & 69.7 $\pm$ 2.7 & 70.5 $\pm$ 5.4 & 74.4 $\pm$ 0.9 & 63.0 $\pm$ 6.1 & 64.7 $\pm$ 2.8 & 75.7 $\pm$ 2.8 & 76.6 $\pm$ 2.9 & 75.3 $\pm$ 2.9 \\
spambase         & 94.7 $\pm$ 0.8 & 98.7 $\pm$ 0.5 & 98.4 $\pm$ 0.4 & 98.0 $\pm$ 0.9 & 98.5 $\pm$ 0.5 & 97.5 $\pm$ 0.8 & 98.3 $\pm$ 0.4 & 98.3 $\pm$ 0.3 & 98.3 $\pm$ 0.5 & 98.6 $\pm$ 0.3 \\
jasmine          & 84.6 $\pm$ 1.7 & 86.2 $\pm$ 0.8 & 85.1 $\pm$ 1.5 & 85.6 $\pm$ 1.3 & 85.3 $\pm$ 1.5 & 81.6 $\pm$ 1.7 & 84.7 $\pm$ 1.7 & 85.9 $\pm$ 1.3 & 86.2 $\pm$ 1.2 & 86.3 $\pm$ 2.0 \\
seismicbumps     & 74.9 $\pm$ 6.8 & 75.6 $\pm$ 8.4 & 73.5 $\pm$ 2.8 & 69.9 $\pm$ 7.4 & 75.1 $\pm$ 9.6 & 70.1 $\pm$ 5.1 & 68.1 $\pm$ 8.4 & 74.3 $\pm$ 6.1 & 75.8 $\pm$ 6.0 & 75.8 $\pm$ 6.2 \\
qsar\_bio        & 84.7 $\pm$ 3.7 & 91.3 $\pm$ 3.1 & 91.0 $\pm$ 3.7 & 91.6 $\pm$ 3.6 & 91.8 $\pm$ 3.8 & 86.0 $\pm$ 3.8 & 91.4 $\pm$ 2.8 & 92.4 $\pm$ 2.5 & 92.4 $\pm$ 2.4 & 92.9 $\pm$ 2.3 \\
\bottomrule
\end{tabular}}
\end{table*}

\begin{table*}
  \caption{Mean percent AUROC scores for the new models under three scenarios. 1) Same as reported above, using leaky gates and MLP skip layers, 2) removing leaky gates but keeping MLP skip layers, and 3) removing both leaky gates and MLP skip layers.}
	\label{tab:ablation-all-scores}
	\centering
		\begin{tabular}{r|rrr|rrr|rrr}
			\toprule
                     & \multicolumn{3}{c}{MLP+} & \multicolumn{3}{c}{PNN}  & \multicolumn{3}{c}{AutoInt}  \\
                     &   skip=T &   skip=T &   skip=F &   skip=T &   skip=T &   skip=F &   skip=T &   skip=T &   skip=F  \\
    Dataset          &   gate=T &   gate=F &   gate=F &   gate=T &   gate=F &   gate=F &   gate=T &   gate=F &   gate=F  \\
    			\midrule
              albert & \bf 74.1 &     73.9 &     73.9 & \bf 74.1 &     74.0 &     73.9 & \bf 73.9 &     73.4 &     73.4  \\
          hcdr\_main & \bf 75.4 &     75.0 &     74.7 & \bf 75.1 &     74.9 &     74.8 & \bf 75.6 &     74.6 &     74.8  \\
          dota2games & \bf 63.4 &     63.4 &     63.3 &     63.3 & \bf 63.4 & \bf 63.4 & \bf 63.4 & \bf 63.4 &     63.2  \\
     bank\_marketing & \bf 93.7 &     93.4 &     93.5 & \bf 93.5 &     93.4 & \bf 93.5 & \bf 93.8 &     93.4 &     93.5  \\
               adult & \bf 76.1 &     75.6 &     76.0 &     75.8 &     75.7 & \bf 75.9 & \bf 76.2 &     75.7 &     76.0  \\
        1995\_income & \bf 91.5 &     91.3 &     91.3 &     91.3 & \bf 91.4 & \bf 91.4 & \bf 91.9 &     91.4 &     91.3  \\
    online\_shoppers & \bf 92.6 &     92.3 &     92.2 & \bf 92.7 &     91.7 &     92.6 & \bf 92.9 &     92.5 &     92.7  \\
            shrutime & \bf 86.2 &     86.0 &     85.8 & \bf 86.4 &     86.0 &     86.1 & \bf 86.6 &     85.8 &     85.5  \\
           blastchar & \bf 84.7 &     84.6 &     84.3 & \bf 84.7 &     84.6 &     84.4 & \bf 84.7 &     84.6 &     84.4  \\
          philippine & \bf 80.7 &     80.3 &     80.2 & \bf 80.8 &     79.8 &     80.4 & \bf 81.1 &     79.1 &     79.3  \\
       insurance\_co & \bf 75.7 &     74.6 &     73.7 & \bf 76.6 &     73.2 &     75.4 & \bf 75.3 &     73.4 &     75.1  \\
            spambase & \bf 98.3 &     98.0 &     97.9 & \bf 98.3 &     97.7 &     97.8 & \bf 98.6 &     98.2 &     98.2  \\
             jasmine &     85.9 & \bf 86.2 &     85.7 & \bf 86.2 &     86.0 &     85.9 & \bf 86.3 &     85.1 &     85.5  \\
        seismicbumps &     74.3 & \bf 74.9 &     72.7 & \bf 75.8 &     75.0 &     75.7 & \bf 75.8 &     75.2 &     74.3  \\
           qsar\_bio &     92.4 & \bf 92.8 &     92.6 &     92.4 &     92.3 & \bf 92.6 & \bf 92.9 &     92.6 &     92.3  \\
            \midrule
               Mean  &     83.0 &     82.8 &     82.5 &     83.1 &     82.6 &     82.9 &     83.3 &     82.6 &     82.6  \\
Number best or tied  &       12 &        3 &        0 &       11 &        2 &        5 &       15 &        1 &        0  \\
			\bottomrule
	\end{tabular}
\end{table*}

\begin{table*}
\caption{Dataset URLs. From \cite{huang2020tabtransformer}.}
\label{tab:dataset-urls}
\centering
\scalebox{0.95}{
\begin{tabular}{ll}
\toprule
     Dataset Name &  URL \\
\midrule
     1995\_income & \url{https://www.kaggle.com/lodetomasi1995/income-classification} \\
            adult & \url{http://automl.chalearn.org/data} \\
           albert & \url{http://automl.chalearn.org/data} \\
  bank\_marketing & \url{https://archive.ics.uci.edu/ml/datasets/bank+marketing} \\
        blastchar & \url{https://www.kaggle.com/blastchar/telco-customer-churn} \\
       dota2games & \url{https://archive.ics.uci.edu/ml/datasets/Dota2+Games+Results} \\
       hcdr\_main & \url{https://www.kaggle.com/c/home-credit-default-risk} \\
    insurance\_co & \url{https://archive.ics.uci.edu/ml/datasets/Insurance+Company+Benchmark+(COIL+2000)} \\
          jasmine & \url{http://automl.chalearn.org/data} \\
 online\_shoppers & \url{https://archive.ics.uci.edu/ml/datasets/Online+Shoppers+Purchasing+Intention+Dataset} \\
       philippine & \url{http://automl.chalearn.org/data} \\
        qsar\_bio & \url{https://archive.ics.uci.edu/ml/datasets/QSAR+biodegradation} \\
     seismicbumps & \url{https://archive.ics.uci.edu/ml/datasets/seismic-bumps} \\
         shrutime & \url{https://www.kaggle.com/shrutimechlearn/churn-modelling} \\
         spambase & \url{https://archive.ics.uci.edu/ml/datasets/Spambase} \\
\bottomrule
\end{tabular}
}
\end{table*}

\end{document}